\documentclass[conference,letterpaper]{IEEEtran}
\IEEEoverridecommandlockouts
\usepackage{cite}
\usepackage{comment}
\usepackage{amsmath,amssymb,amsfonts}
\usepackage{algorithmic}
\usepackage{graphicx}
\usepackage{textcomp}
\usepackage{xcolor}
\usepackage{pgfplots}
\usepackage{pgfplotstable}
\usepackage{subcaption}
\pgfplotsset{compat=1.18}

\usepackage{cite}
\usepackage{comment}
\usepackage{amsmath,amssymb,amsfonts}
\usepackage{algorithmic}
\usepackage{graphicx}
\usepackage{textcomp}
\usepackage{xcolor}
\usepackage{pgfplots}
\usepackage{pgfplotstable}
\usepackage{subcaption}
\usepackage{tikz}
\usepackage[utf8]{inputenc}
\usepackage[T1]{fontenc}
\usepackage{caption}
\usepackage{helvet}
\usetikzlibrary{shapes, arrows.meta, positioning, fit, shadows, backgrounds, patterns, decorations.pathmorphing, shapes,  calc}
\DeclareMathOperator{\Tr}{Tr}

\definecolor{primaryBlue}{RGB}{41, 128, 185}
\definecolor{accentOrange}{RGB}{230, 126, 34}
\definecolor{softGray}{RGB}{236, 240, 241}
\definecolor{darkGray}{RGB}{52, 73, 94}
\definecolor{emeraldGreen}{RGB}{46, 204, 113}
\definecolor{softRed}{RGB}{231, 76, 60}
\definecolor{gold}{RGB}{255, 215, 0}

\definecolor{oceanBlue}{RGB}{41, 128, 185}
\definecolor{deepPurple}{RGB}{142, 68, 173}
\definecolor{emeraldGreen}{RGB}{46, 204, 113}
\definecolor{darkSlate}{RGB}{44, 62, 80}
\definecolor{coral}{RGB}{231, 76, 60}
\definecolor{silver}{RGB}{189, 195, 199}
\definecolor{cloudWhite}{RGB}{236, 240, 241}
\definecolor{midnightBlue}{RGB}{44, 62, 80}
\definecolor{lightBlue}{RGB}{52, 152, 219}

\definecolor{gradStart}{RGB}{41, 128, 185}
\definecolor{gradEnd}{RGB}{52, 152, 219}
\definecolor{purpleStart}{RGB}{142, 68, 173}
\definecolor{purpleEnd}{RGB}{155, 89, 182}
\definecolor{greenStart}{RGB}{39, 174, 96}
\definecolor{greenEnd}{RGB}{46, 204, 113}

\definecolor{featureRed}{RGB}{180, 60, 60}
\definecolor{featureRedDark}{RGB}{140, 40, 40}
\definecolor{featureRedLight}{RGB}{200, 100, 100}
\definecolor{featureGreen}{RGB}{120, 160, 120}
\definecolor{featureGreenDark}{RGB}{80, 120, 80}
\definecolor{featureGreenLight}{RGB}{140, 180, 140}
\definecolor{blockPurple}{RGB}{200, 180, 220}
\definecolor{blockYellow}{RGB}{250, 230, 150}
\definecolor{snrOrange}{RGB}{255, 180, 100}

\def\BibTeX{{\rm B\kern-.05em{\sc i\kern-.025em b}\kern-.08em
    T\kern-.1667em\lower.7ex\hbox{E}\kern-.125emX}}
\begin{document}

\title{Doubly Adaptive Channel and Spatial Attention for Semantic Image Communication in IoT Networks \\} 

\author{
\IEEEauthorblockN{Soroosh Miri \IEEEauthorrefmark{1}, Sepehr Abolhasani \IEEEauthorrefmark{1}, Shahrokh Farahmand \IEEEauthorrefmark{1}, S. Mohammad Razavizadeh \IEEEauthorrefmark{1}, and Jiguang He \IEEEauthorrefmark{2}}
\IEEEauthorblockA{
\IEEEauthorrefmark{1} \textit{School of Electrical Engineering,
Iran University of Science and Technology (IUST),
Tehran, Iran} \\
\IEEEauthorrefmark{2} \textit{Great Bay
University, Dongguan 523000, China}
}}


\maketitle

\begin{abstract}
Internet of Things (IoT) networks face significant challenges such as limited communication bandwidth, constrained computational and energy resources, and highly dynamic wireless channel conditions. Utilization of deep neural networks (DNNs) combined with semantic communication has emerged as a promising paradigm to address these limitations. Deep joint source-channel coding (DJSCC) has recently been proposed to enable semantic communication of images. Building upon the original DJSCC formulation, low-complexity attention-style architectures have been added to the DNNs for further performance enhancement. As a main hurdle, training these DNNs separately for various signal-to-noise ratios (SNRs) will amount to excessive storage or communication overhead, which can not be maintained by small IoT devices. SNR adaptive DJSCC (ADJSCC), has been proposed to train the DNNs once but feeds the current SNR as part of the data to the channel-wise attention mechanism. We improve upon ADJSCC by a simultaneous utilization of doubly adaptive channel-wise and spatial attention modules at both transmitter and receiver. These modules dynamically adjust to varying channel conditions and spatial feature importance, enabling robust and efficient feature extraction and semantic information recovery. Simulation results corroborate that our proposed doubly adaptive DJSCC (DA-DJSCC) significantly improves upon ADJSCC in several performance criteria, while incurring a mild increase in complexity. These facts render DA-DJSCC a desirable choice for semantic communication in performance demanding but low-complexity IoT networks.
\end{abstract}




\section{Introduction}
During the last two decades, IoT-relevant technologies have been advanced rapidly, enabling large-scale interconnection of physical objects for intelligent sensing and communication. These developments have enhanced big data analytics, interoperability, and network efficiency. However, the exponential growth in the number of IoT devices imposes major challenges, including data overload, bandwidth scarcity, and computational and energy resource constraints on small IoT devices \cite{b1}.

Concurrently, a new paradigm known as semantic communication has emerged beyond conventional wireless communications. Unlike traditional approaches focused on accurate bit-level transmission, semantic communication aims to transmit and recover the intended meaning of transmitted data. By shifting the focus from syntax (bits and symbols) to semantics (meaning and context), it enables more efficient, robust, and intelligent information exchange \cite{b2}. Compared to conventional communication systems, semantic communications offers several key advantages. First, by transmitting only the information that are semantically relevant to the receiver’s task, semantic communication can significantly reduce communication overhead and improve spectral and energy efficiency. Second, by focusing on the meaning rather than the exact bit representation, it enhances robustness against channel noise and data loss, ensuring reliable interpretation even under imperfect channel conditions. Third, semantic communication enables intelligent and goal-oriented information exchange, where the system adapts dynamically to the context, intent, and task requirements of the application \cite{b3,b4}. These advantages make semantic communication particularly suitable for resource-constrained and mission-critical IoT scenarios, where efficient and adaptive information exchange is essential \cite{b5}.

Motivated by the introduction of Deep joint source-channel coding (DJSCC) in \cite{b6}, numerous recent studies have focused on semantic image transmission. Several works have enhanced DJSCC performance by incorporating attention mechanisms into the encoder–decoder architecture \cite{b7,b8,b9}. Subsequently, channel-wise and spatial attention modules were added into the DJSCC framework in \cite{b10}. Reference \cite{b11} further extended \cite{b10} to the IoT setup, where both low-complexity spatial and channel-wise attention mechanisms tailored to IoT devices were proposed. Nevertheless, the approach in \cite{b11} lacks explicit channel adaptivity in both the spatial and channel attention modules. The SNR adaptivity for a channel-wise attention mechanism has been addressed by \cite{b7}, which is referred to by Adaptive DJSCC (ADJSCC). However, simultaneous incorporation of double adaptivity in both spatial and channel-wise attention modules for image transmission is missing from the literature. To address this gap, our proposed architecture introduces doubly adaptive attention modules that dynamically adjust to varying SNR, channel-wise and spatial conditions, enhancing robustness and efficiency in IoT environments.


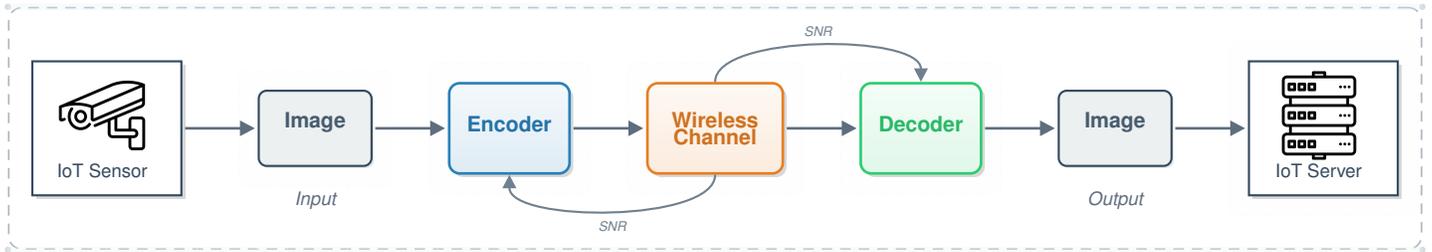
\begin{figure*}[t!]
\centering
\hspace*{-0.7cm}
\resizebox{\textwidth}{!}{%
\begin{tikzpicture}[
    scale=0.7,
    node distance=1.5cm,
    sensor_box/.style={
        rectangle, 
        draw=darkGray, 
        minimum width=1.8cm, 
        minimum height=1.5cm,
        align=center, 
        fill=white,
        line width=0.8pt,
        drop shadow={shadow xshift=0.5mm, shadow yshift=-0.5mm, opacity=0.3}
    },
    image_box/.style={
        rectangle, 
        rounded corners=3pt,
        draw=darkGray, 
        minimum width=1.5cm, 
        minimum height=1cm,
        align=center, 
        fill=softGray,
        line width=0.8pt,
        font=\footnotesize\sffamily\bfseries,
        drop shadow={shadow xshift=0.3mm, shadow yshift=-0.3mm, opacity=0.2}
    },
    encoder_box/.style={
        rectangle,
        rounded corners=4pt,
        draw=primaryBlue,
        minimum width=1.6cm,
        minimum height=1.2cm,
        align=center,
        top color=primaryBlue!5,
        bottom color=primaryBlue!15,
        line width=1pt,
        font=\footnotesize\sffamily,
        text=primaryBlue!90!black,
        drop shadow={shadow xshift=0.5mm, shadow yshift=-0.5mm, opacity=0.3}
    },
    channel_box/.style={
        rectangle,
        rounded corners=4pt,
        draw=accentOrange,
        minimum width=1.8cm,
        minimum height=1.2cm,
        align=center,
        top color=accentOrange!5,
        bottom color=accentOrange!15,
        line width=1pt,
        font=\footnotesize\sffamily\bfseries,
        text=accentOrange!90!black,
        drop shadow={shadow xshift=0.5mm, shadow yshift=-0.5mm, opacity=0.3}
    },
    decoder_box/.style={
        rectangle,
        rounded corners=4pt,
        draw=emeraldGreen,
        minimum width=1.6cm,
        minimum height=1.2cm,
        align=center,
        top color=emeraldGreen!5,
        bottom color=emeraldGreen!15,
        line width=1pt,
        font=\footnotesize\sffamily,
        text=emeraldGreen!90!black,
        drop shadow={shadow xshift=0.5mm, shadow yshift=-0.5mm, opacity=0.3}
    },
    server_box/.style={
        rectangle, 
        draw=darkGray, 
        minimum width=1.8cm, 
        minimum height=1.5cm,
        align=center, 
        fill=white,
        line width=0.8pt,
        drop shadow={shadow xshift=0.5mm, shadow yshift=-0.5mm, opacity=0.3}
    },
    arrow/.style={
        -{Triangle[length=2mm, width=2mm]},
        line width=1pt,
        color=darkGray!80,
        shorten >=1pt,
        shorten <=1pt
    },
    feedback_arrow/.style={
        -{Triangle[length=1.8mm, width=1.8mm]},
        line width=0.8pt,
        darkGray!70
    },
    label_style/.style={
        font=\scriptsize\sffamily\itshape,
        text=darkGray!80
    },
    section_frame/.style={
        rectangle,
        draw=darkGray!30,
        dashed,
        dash pattern=on 3pt off 2pt,
        line width=0.4pt,
        rounded corners=3pt
    }
]

\node[sensor_box] (sensor) {
    \begin{tabular}{c}
    \includegraphics[width=1.2cm]{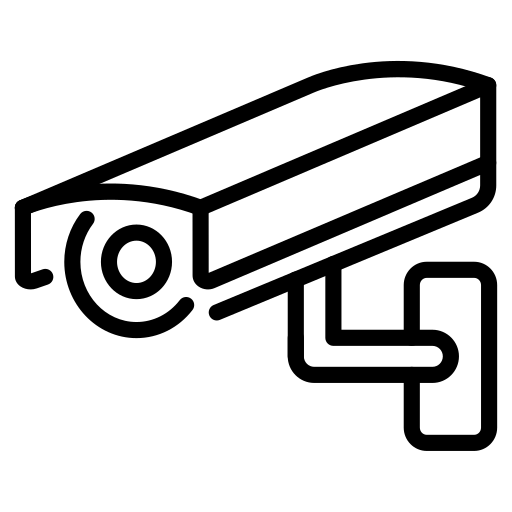}\\[-2mm]
    \scriptsize\sffamily\color{darkGray} IoT Sensor
    \end{tabular}
};

\node[image_box, right=1cm of sensor] (image1) {
    \color{darkGray!90}Image\\[-1mm]\color{darkGray!90}
};

\node[encoder_box, right=1cm of image1] (encoder) {
    \textbf{Encoder}\\[-2mm]
    
};

\node[channel_box, right=1cm of encoder] (channel) {
    Wireless\\Channel
};

\node[decoder_box, right=1cm of channel] (decoder) {
    \textbf{Decoder}\\[-2mm]
    
};

\node[image_box, right=1cm of decoder] (image2) {
    \color{darkGray!90}Image\\[-1mm]\color{darkGray!90}
};

\node[server_box, right=1cm of image2] (server) {
    \begin{tabular}{c}
    \includegraphics[width=1.2cm]{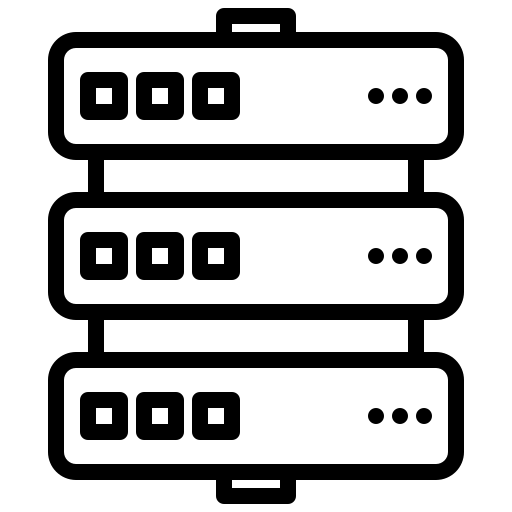}\\[-2mm]
    \scriptsize\sffamily\color{darkGray} IoT Server
    \end{tabular}
};

\draw[arrow] (sensor) -- (image1);
\draw[arrow] (image1) -- (encoder);
\draw[arrow] (encoder) -- (channel);
\draw[arrow] (channel) -- (decoder);
\draw[arrow] (decoder) -- (image2);
\draw[arrow] (image2) -- (server);

\draw[feedback_arrow] 
    (channel.south) .. controls +(0,-0.8) and +(0,-0.8) .. 
    node[pos=0.5, below, font=\tiny\sffamily\itshape, text=darkGray!70] {SNR} 
    (encoder.south);

\draw[feedback_arrow] 
    (channel.north) .. controls +(0,0.8) and +(0,0.8) .. 
    node[pos=0.5, above, font=\tiny\sffamily\itshape, text=darkGray!70] {SNR} 
    (decoder.north);

\node[label_style, below=2mm of image1] {Input};
\node[label_style, below=2mm of image2] {Output};

\begin{scope}[on background layer]
    \node[section_frame, fill=softGray!20, fill opacity=0.3, fit=(image1), inner sep=2.5mm] {};
    \node[section_frame, fill=primaryBlue!5, fill opacity=0.5, fit=(encoder), inner sep=2.5mm] {};

    \node[section_frame, fill=accentOrange!5, fill opacity=0.5, fit=(channel), inner sep=2.5mm] {};

    \node[section_frame, fill=emeraldGreen!5, fill opacity=0.5, fit=(decoder), inner sep=2.5mm] {};
    \node[section_frame, fill=darkGray!5, fill opacity=0.5, fit=(server), inner sep=2.5mm] {};

    \node[
        rectangle,
        draw=darkGray!40,
        dashed,
        dash pattern=on 4pt off 3pt,
        line width=0.6pt,
        rounded corners=5pt,
        fit=(sensor) (image1) (encoder) (channel) (decoder) (image2) (server),
        inner xsep=3mm,
        inner ysep=7mm,
        fill=white,
        fill opacity=0.95
    ] (systemframe) {};

    \foreach \corner in {north west, north east, south west, south east} {
        \fill[darkGray!20] (systemframe.\corner) circle (1.5pt);
    }

    \fill[primaryBlue!10, opacity=0.3] 
        ([yshift=-0.5mm]systemframe.north west) rectangle ([yshift=0.5mm]systemframe.north east);
    \fill[emeraldGreen!10, opacity=0.3] 
        ([yshift=-0.5mm]systemframe.south west) rectangle ([yshift=0.5mm]systemframe.south east);
\end{scope}

\begin{scope}[on background layer]
    \foreach \x in {0,0.5,...,10} {
        \foreach \y in {0,0.5,...,3} {
            \fill[darkGray!5] (\x,-\y) circle (0.3pt);
        }
    }
\end{scope}

\end{tikzpicture}
}
\caption{System model of the proposed DA-DJSCC approach.}
     \label{fig:system_model}
\end{figure*}

\section{System Model and Problem Formulation}

We consider a point-to-point wireless link between a single antenna IoT device and a single antenna server as shown in Fig.~\ref{fig:system_model}. The wireless channel is modeled by a multiplicative fading gain. Afterwards, white Gaussian noise (WGN) is added to the channel output at the receiver. Assuming that the transmitted feature matrix by the transmitter is represented by $\mathbf{X}$, and the received matrix at the receiver is denoted by $\mathbf{Y}$, the following relation holds
\begin{equation} \label{fading}
    \mathbf{Y}=h\mathbf{X}+\mathbf{N},
\end{equation}
where $h$ denotes the perfectly known complex channel gain between the single-antenna transmitter and receiver. Upon using a Zero-forcing (ZF) equalizer, both sides of \eqref{fading} are multiplied by $(h^Hh)^{-1}h^{H}=h^*/|h|^2$. We obtain
\begin{equation} \label{awgn}
    \frac{h^{*}\mathbf{Y}}{|h|^2}=\mathbf{X}+\frac{h^{*}\mathbf{N}}{|h|^2},
\end{equation}
which can be rewritten as
\begin{equation} \label{awgn2}
    \tilde{\mathbf{Y}}=\mathbf{X}+\tilde{\mathbf{N}}.
\end{equation}
This amounts to adding scaled WGN to the channel output. The only effect that the known fading gain has is to modify the received signal-to-noise ratio (SNR). Assuming that the transmitted $\mathbf{X}$ is scaled to have a unit Frobenius norm, the SNR is given by 
\begin{equation}
    {\rm SNR}:=\frac{|h|^2\Tr(\mathbf{X}^H\mathbf{X})}{\mathbb{E}\left[\Tr(\mathbf{N}^H\mathbf{N})\right]}=\frac{|h|^2}{k\sigma_n^2},
\end{equation}
where $\sigma_n^2$ is the power of any scalar noise element and $k$ denotes the size of $\mathbf{X}$, which is the number of rows times the number of columns. It should be mentioned that $\mathbb{E}\left[\Tr(\mathbf{N}^H\mathbf{N})\right]=\mathbb{E}\left[\|\mathbf{N}\|_F^2\right]$, which is the expected value of Frobenius norm squared. This SNR definition and the corresponding definition for $k$ can be easily extended to the case where $\mathbf{X}$ is a tensor. This happens when processed feature maps over several channels are transmitted through the fading channel.

As depicted in Fig. 1, the data captured by an IoT device, which is an image in this work, is first processed by a semantic encoder. The encoder extracts and encodes the most important semantic features from the input data while adapting to the current SNR or channel conditions. The extracted semantic features are then transmitted over a wireless channel, which can be modeled as \eqref{fading} with known fading gain. At the receiver side, a semantic decoder reconstructs the received semantic representations into the original image content. The reconstructed image or feature map at the receiver side is subsequently transmitted to an edge or cloud server, where a recognition or inference task, such as image classification or object detection, is performed. The goal of our design is to maximize the semantic content of the reconstructed image at the receiver, such that the downstream machine learning task can enjoy an improved accuracy. In designing the corresponding transmitter and receiver, several competing objectives must be addressed simultaneously. They include minimizing transmission bandwidth, and power consumption, ensuring reliable transmission over noisy channels, and maximizing inference task accuracy at the downstream machine learning server. Overall, the design of such IoT communications links require joint consideration of four key aspects: (i) data compression to reduce bandwidth usage and energy cost, (ii) channel coding to improve transmission reliability, (iii) improving the reconstruction accuracy at the receiver, and (iv) respecting the limited computational capability of IoT devices \cite{b12}. 

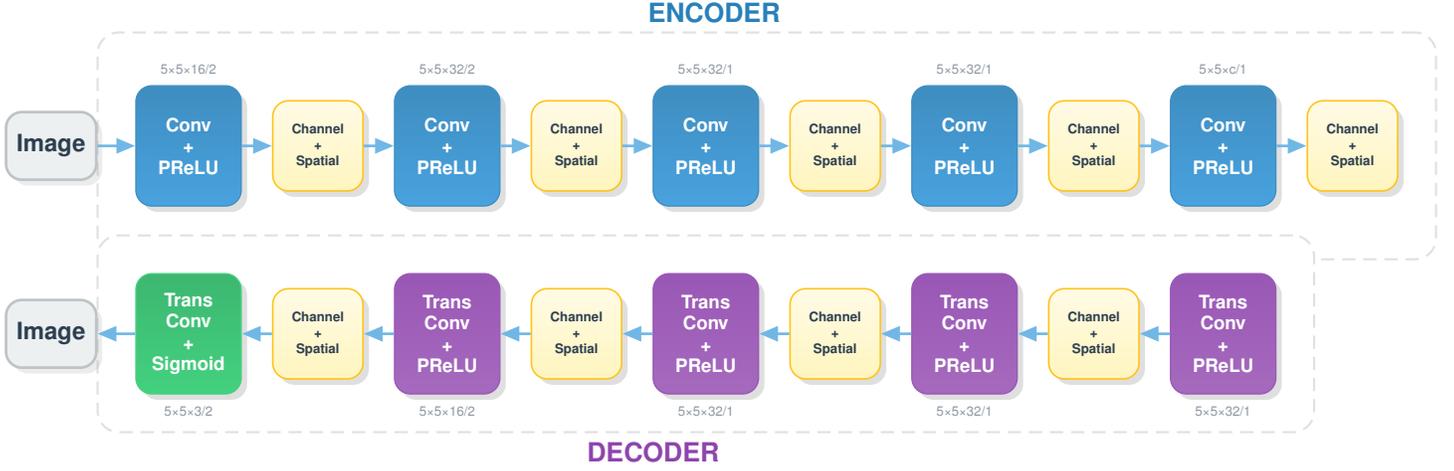
\begin{figure*}[t!]
\centering
\resizebox{\textwidth}{!}{%
\begin{tikzpicture}[
    scale=0.35,
    node distance=0.45cm,
    io_block/.style={
        rectangle,
        rounded corners=6pt,
        draw=silver,
        fill=cloudWhite,
        minimum width=1.2cm,
        minimum height=0.9cm,
        font=\sffamily\small\bfseries,
        text=darkSlate,
        line width=1pt,
        drop shadow={opacity=0.15}
    },
    conv_block/.style={
        rectangle,
        rounded corners=6pt,
        draw=oceanBlue!90,
        top color=gradStart!90,
        bottom color=gradEnd!90,
        minimum width=1.4cm,
        minimum height=1.6cm,
        text=white,
        font=\sffamily\scriptsize\bfseries,
        align=center,
        line width=0.5pt,
        drop shadow={opacity=0.25}
    },
    attention_block/.style={
        rectangle,
        rounded corners=6pt,
        draw=gold!70!orange!90,
        top color=gold!10,
        bottom color=gold!25,
        minimum width=1.2cm,
        minimum height=1.2cm,
        align=center,
        font=\sffamily\tiny\bfseries,
        text=darkSlate,
        line width=0.6pt,
        drop shadow={opacity=0.25}
    },
    trans_conv_block/.style={
        rectangle,
        rounded corners=6pt,
        draw=deepPurple!90,
        top color=purpleStart!90,
        bottom color=purpleEnd!90,
        minimum width=1.4cm,
        minimum height=1.6cm,
        text=white,
        font=\sffamily\scriptsize\bfseries,
        align=center,
        line width=0.5pt,
        drop shadow={opacity=0.25}
    },
    final_conv_block/.style={
        rectangle,
        rounded corners=6pt,
        draw=emeraldGreen!90,
        top color=greenStart!90,
        bottom color=greenEnd!90,
        minimum width=1.4cm,
        minimum height=1.6cm,
        text=white,
        font=\sffamily\scriptsize\bfseries,
        align=center,
        line width=0.5pt,
        drop shadow={opacity=0.25}
    },
    channel_block/.style={
        rectangle,
        rounded corners=10pt,
        draw=coral!80,
        top color=coral!85,
        bottom color=coral!95,
        minimum width=1.5cm,
        minimum height=5.5cm,
        text=white,
        font=\sffamily\small\bfseries,  
        align=center,
        line width=0.8pt,
        drop shadow={opacity=0.35},
        pattern=north east lines,
        pattern color=white!10
    },
    frame_style/.style={
        rectangle,
        draw=silver!50,
        rounded corners=8pt,
        line width=0.8pt,
        fill=white,
        fill opacity=0.97,
        dash pattern=on 5pt off 3pt
    },
    arrow/.style={
        -{Triangle[length=2.5mm, width=2mm]},
        line width=1pt,
        lightBlue!70
    },
    param_label/.style={
        font=\tiny\sffamily,
        text=midnightBlue!60,
        fill=white,
        fill opacity=0.9,
        rounded corners=2pt,
        inner sep=2pt
    }
]

\node[io_block] (input) at (-9, 2.5) {Image};

\node[conv_block, right=0.5cm of input] (conv1) {Conv\\+\\PReLU};
\node[param_label, above=2pt of conv1] {5×5×16/2};

\node[attention_block, right=0.4cm of conv1] (att1) {Channel\\+\\Spatial};

\node[conv_block, right=0.4cm of att1] (conv2) {Conv\\+\\PReLU};
\node[param_label, above=2pt of conv2] {5×5×32/2};

\node[attention_block, right=0.4cm of conv2] (att2) {Channel\\+\\Spatial};

\node[conv_block, right=0.4cm of att2] (conv3) {Conv\\+\\PReLU};
\node[param_label, above=2pt of conv3] {5×5×32/1};

\node[attention_block, right=0.4cm of conv3] (att3) {Channel\\+\\Spatial};

\node[conv_block, right=0.4cm of att3] (conv4) {Conv\\+\\PReLU};
\node[param_label, above=2pt of conv4] {5×5×32/1};

\node[attention_block, right=0.4cm of conv4] (att4) {Channel\\+\\Spatial};

\node[conv_block, right=0.4cm of att4] (conv5) {Conv\\+\\PReLU};
\node[param_label, above=2pt of conv5] {5×5×c/1};

\node[attention_block, right=0.4cm of conv5] (att5) {Channel\\+\\Spatial};

\node[io_block] (output) at (-9, -2.5) {Image};

\node[final_conv_block] (dconv1) at (conv1 |- output) {Trans\\Conv\\+\\Sigmoid};
\node[param_label, below=2pt of dconv1] {5×5×3/2};

\node[attention_block, right=0.4cm of dconv1] (datt1) {Channel\\+\\Spatial};

\node[trans_conv_block, right=0.4cm of datt1] (dconv2) {Trans\\Conv\\+\\PReLU};
\node[param_label, below=2pt of dconv2] {5×5×16/2};

\node[attention_block, right=0.4cm of dconv2] (datt2) {Channel\\+\\Spatial};

\node[trans_conv_block, right=0.4cm of datt2] (dconv3) {Trans\\Conv\\+\\PReLU};
\node[param_label, below=2pt of dconv3] {5×5×32/1};

\node[attention_block, right=0.4cm of dconv3] (datt3) {Channel\\+\\Spatial};

\node[trans_conv_block, right=0.4cm of datt3] (dconv4) {Trans\\Conv\\+\\PReLU};
\node[param_label, below=2pt of dconv4] {5×5×32/1};

\node[attention_block, right=0.4cm of dconv4] (datt4) {Channel\\+\\Spatial};

\node[trans_conv_block, right=0.4cm of datt4] (dconv5) {Trans\\Conv\\+\\PReLU};
\node[param_label, below=2pt of dconv5] {5×5×32/1};


\draw[arrow] (input) -- (conv1);
\draw[arrow] (conv1) -- (att1);
\draw[arrow] (att1) -- (conv2);
\draw[arrow] (conv2) -- (att2);
\draw[arrow] (att2) -- (conv3);
\draw[arrow] (conv3) -- (att3);
\draw[arrow] (att3) -- (conv4);
\draw[arrow] (conv4) -- (att4);
\draw[arrow] (att4) -- (conv5);
\draw[arrow] (conv5) -- (att5);

\draw[arrow] (dconv5) -- (datt4);
\draw[arrow] (datt4) -- (dconv4);
\draw[arrow] (dconv4) -- (datt3);
\draw[arrow] (datt3) -- (dconv3);
\draw[arrow] (dconv3) -- (datt2);
\draw[arrow] (datt2) -- (dconv2);
\draw[arrow] (dconv2) -- (datt1);
\draw[arrow] (datt1) -- (dconv1);
\draw[arrow] (dconv1) -- (output);

\end{tikzpicture}
}
\caption{Overall architecture of the proposed DA-DJSCC model.}
\label{fig:architecture}
\end{figure*}

\section{Proposed Transceiver Design}
As mentioned before, simultaneous channel-wise and spatial attention modules have been utilized to improve the semantic reconstruction capability at the receiver. For time-varying wireless channels, the SNR changes quickly. Hence, at every SNR a different set of DNN weights should be utilized. This significantly increases the training time, even if it is performed offline, and demands ample storage and communication capability to store and communicate the weights. The recent remedy for time-varying channels is to aggregate the input feature map with the current SNR value. As a result, as channel SNR varies over time, we can utilize the same set of weights, but feed the changing SNR to the transmitter and receiver DNNs as part of the input. As shown in Fig.~\ref{fig:architecture}, the overall model comprises a semantic encoder and a semantic decoder. These two components are jointly optimized in an end-to-end manner to maximize the semantic performance rather than conventional bit-level reconstruction accuracy. To ensure adaptive and robust communication, the model incorporates both channel-wise and spatial attention modules within the encoder–decoder architecture, enabling dynamic adjustment to changes in channel conditions and feature importance. Finally, it should be noted that our implemented attention modules enjoy a significantly lower complexity compared to standard attention mechanisms. This design choice makes the modified attention blocks ideal options for implementation in low-complexity IoT devices. We refer to our architecture and the corresponding algorithm as doubly adaptive deep joint source and channel coding (DA-DJSCC).

\subsection{Semantic Encoder Design}\label{AA} 

The semantic encoder is responsible for extracting and transmitting semantic features from the input image. As shown in Fig.~\ref{fig:architecture}, the encoder, which is represented by the top row blocks, is composed of three main components: a feature extractor, a channel-wise attention module, and a spatial attention module.
Let $\mathbf{S} \in \mathbb{R}^{H \times W \times C}$ denote the input color image captured by an IoT device. Here, $H,W$ and $C$ denote the number of pixels in the height and width respectively, while $C$ denote the number of color channels. We can set $C=1$ for gray-scale images. The image is first passed through a convolutional neural network (CNN)-based feature extractor, which produces an initial feature map:
\begin{equation}
    \mathbf{F}_0 = \mathbf{G}(\mathbf{S}; \boldsymbol{\theta}_g^0), 
\end{equation}
where $\mathbf{G}(\cdot)$ represents the CNN-based feature extraction function with trainable parameters $\boldsymbol{\theta}_g^0$.

The extracted feature map $\mathbf{F}_0$ is then processed through a series of  attention-enhanced encoding modules. For the $\ell$-th module $(\ell = 1, 2, \ldots, L)$, the feature transformation can be expressed as:
\begin{equation}
    \mathbf{F}_{\ell} = \mathbf{A}_s\left(\mathbf{A}_c\left(\mathbf{G}\left(\mathbf{F}_{\ell-1}; \boldsymbol{\theta}_g^\ell \right); \boldsymbol{\theta}_c^\ell \right); \boldsymbol{\theta}_s^\ell \right),
\end{equation}
where $\mathbf{G}(\cdot)$, $\mathbf{A}_c(\cdot)$ and $\mathbf{A}_s(\cdot)$ denote the convolutional layer, channel and spatial attention functions, respectively, and $\boldsymbol{\theta}_g^\ell$, $\boldsymbol{\theta}_c^\ell$, and $\boldsymbol{\theta}_s^\ell$ represent the trainable parameters of the corresponding modules.
After passing through all $L$ modules, the final encoded semantic feature representation is obtained as:
\begin{equation}
    \mathbf{X} = \mathbf{F}_L \quad , \quad \mathbf{X} \in\mathbb{R}^{H_L\times W_L \times C_L}
\end{equation}
 Where $k:=H_L W_L C_L$ represents the number of transmitted symbols after joint source and channel coding. By stacking multiple feature–attention modules, the encoder progressively refines its representation, focusing on the most informative semantic features. Moreover, during training, the attention weights are conditioned on the SNR, allowing the encoder to adapt its feature emphasis dynamically for improved robustness to channel variations. The compression ratio for the encoder is defined as $k/n$, where $k$ is defined above and $n:=H_0 W_0 C_0$.

\subsection{Semantic Decoder}

The semantic decoder is represented by the bottom row blocks in Fig. \ref{fig:architecture}. It reconstructs the original image from the feature representation received through the wireless channel.  Let $\tilde{\mathbf{Y}} \in \mathbb{R}^{H_L \times W_L}$ denote the received feature vector as in \eqref{awgn2}. The decoder reconstructs the image through a sequence of operations in each of its $L$ stacked modules. In the first block at the receiver, $\tilde{\mathbf{Y}}$ is first processed by a transposed convolutional layer $\mathbf{T}(\cdot)$ to upsample and refine the feature representation. The output of this block is then passed through a cascade of channel attention module, $\mathbf{A}_c(\cdot)$, and spatial attention module $\mathbf{A}_s(\cdot)$, both of which adaptively adjust their attention weights according to the channel SNR; see Fig. \ref{fig:architecture}. At the output of the $\ell$-th layer we have
\begin{equation}
    \tilde{\mathbf{F}}_\ell = \mathbf{A}_s\left(\mathbf{A}_c\left(\mathbf{T}\left(\tilde{\mathbf{F}}_{\ell-1}; \boldsymbol{\tilde\theta}_g^\ell\right) ;\boldsymbol{\tilde\theta}_c^\ell\right); \boldsymbol{\tilde\theta}_s^\ell\right),
\end{equation}
where $\boldsymbol{\tilde\theta}_c^\ell$ and $\boldsymbol{\tilde\theta}_s^\ell$ denote the learnable parameters of the channel-wise and spatial attention modules, respectively. Furthermore, $\boldsymbol{\tilde\theta}_g^\ell$ denote the learnable parameters of the transposed convolutional layer and we have $\tilde{\mathbf{F}}_0=\tilde{\mathbf{Y}}$.

After passing through all $L$ stacked modules, the final output of the decoder is the reconstructed image $\hat{\mathbf{S}}$
\begin{equation}
    \hat{\mathbf{S}}:=\tilde{\mathbf{F}}_L.
\end{equation}

\subsection{Adaptive Channel and Spatial Attention Modules}

The adaptive channel attention module is designed to emphasize the most informative feature channels according to both the input feature map and the current channel condition. Given an input feature map $\mathbf{F} \in \mathbb{R}^{H \times W \times C}$, where $H$, $W$, and $C$ denote the height, width, and number of channels, respectively, the module adaptively learns a channel-wise weighting vector guided by the SNR.

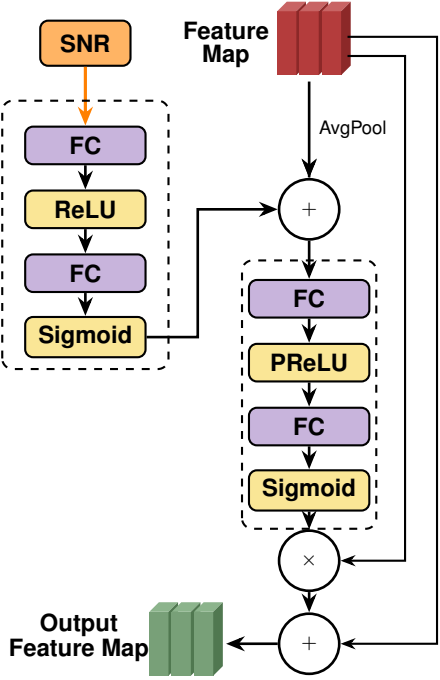
\begin{figure}[t!]
\centering
\resizebox{0.3\textwidth}{!}{
\begin{tikzpicture}[
    scale=0.85,
    process_block/.style={
        rectangle,
        rounded corners=3pt,
        draw=black,
        fill=#1,
        minimum width=1.6cm,
        minimum height=0.5cm,
        font=\sffamily\small\bfseries,
        text=black,
        line width=0.8pt
    },
    snr_block/.style={
        rectangle,
        rounded corners=3pt,
        draw=black,
        fill=snrOrange,
        minimum width=1.2cm,
        minimum height=0.6cm,
        font=\sffamily\small\bfseries,
        text=black,
        line width=0.8pt
    },
    operation/.style={
        circle,
        draw=black,
        fill=white,
        minimum size=0.8cm,
        font=\sffamily\scriptsize,
        line width=1pt
    },
    frame_style/.style={
        rectangle,
        draw=black,
        rounded corners=4pt,
        line width=0.8pt,
        dashed,
        fill=white,
        fill opacity=0
    },
    arrow/.style={
        -{Stealth[length=2.5mm, width=2mm]},
        line width=1pt,
        black
    },
    bypass_arrow/.style={
        -{Stealth[length=2mm, width=1.5mm]},
        line width=0.8pt,
        black
    }
]


\begin{scope}[shift={(2, 0.8)}]
    \foreach \i in {0, 0.35, 0.7} {
        \fill[featureRed] (\i, -0.5) rectangle (\i+0.3, 0.5);
        \fill[featureRedLight] (\i, 0.5) -- (\i+0.3, 0.5) -- (\i+0.45, 0.65) -- (\i+0.15, 0.65) -- cycle;
        \fill[featureRedDark] (\i+0.3, -0.5) -- (\i+0.3, 0.5) -- (\i+0.45, 0.65) -- (\i+0.45, -0.35) -- cycle;
    }
    \node[font=\sffamily\small\bfseries] at (-0.8, 0.2) {Feature};
    \node[font=\sffamily\small\bfseries] at (-0.8, -0.2) {Map};
\end{scope}

\node[snr_block] (snr) at (-1, 0.8) {SNR};

\node[process_block=blockPurple] (fc1) at (-1, -0.8) {FC};
\node[process_block=blockYellow] (relu) at (-1, -1.8) {ReLU};
\node[process_block=blockPurple] (fc2) at (-1, -2.8) {FC};
\node[process_block=blockYellow] (relu2) at (-1, -3.8) {Sigmoid};

\draw[arrow, orange, line width=1.2pt] (snr) -- (fc1);
\draw[arrow] (fc1) -- (relu);
\draw[arrow] (relu) -- (fc2);
\draw[arrow] (fc2) -- (relu2);

\draw[frame_style, minimum width=2.2cm, minimum height=3.8cm] (-2.3, -4.3) rectangle (0.3, -0.1);

\draw[frame_style, minimum width=2.2cm, minimum height=3.8cm] (1.45, -6.8) rectangle (3.55, -2.6);

\draw[arrow] (2.5, 0.2) -- (2.5, -1.3) node[midway, right, font=\sffamily\scriptsize] {AvgPool};

\node[process_block=blockPurple] (add1) at (2.5, -1.8) {Concatenator};

\draw[arrow] (relu2.east) -- ++(0.8, 0) |- (add1);

\node[process_block=blockPurple] (fc3) at (2.5, -3.2) {FC};
\node[process_block=blockYellow] (relu3) at (2.5, -4.2) {PReLU};
\node[process_block=blockPurple] (fc4) at (2.5, -5.2) {FC};
\node[process_block=blockYellow] (relu4) at (2.5, -6.2) {Sigmoid};
\draw[arrow] (add1) -- (fc3);
\draw[arrow] (fc3) -- (relu3);
\draw[arrow] (relu3) -- (fc4);
\draw[arrow] (fc4) -- (relu4);

\node[operation] (mult) at (2.5, -7.3) {$\times$};
\draw[arrow] (relu4) -- (mult);

\draw[bypass_arrow] (3.1, 0.6) -- (4, 0.6) -- (4, -7.3) -- (mult.east);

\node[operation] (add2) at (2.5, -8.6) {$+$};
\draw[arrow] (mult) -- (add2);

\draw[bypass_arrow] (3.1, 0.9) -- (4.5, 0.9) -- (4.5, -8.6) -- (add2.east);

\node[font=\sffamily\small\bfseries] at (-1.7+0.6, -9.8+1.5) {Output};
\node[font=\sffamily\small\bfseries] at (-1.7+0.6, -10.2+1.5) {Feature Map};

\begin{scope}[shift={(0, -8.6)}]
    \foreach \i in {0, 0.35, 0.7} {
        \fill[featureGreen] (\i, -0.5) rectangle (\i+0.3, 0.5);
        \fill[featureGreenLight] (\i, 0.5) -- (\i+0.3, 0.5) -- (\i+0.45, 0.65) -- (\i+0.15, 0.65) -- cycle;
        \fill[featureGreenDark] (\i+0.3, -0.5) -- (\i+0.3, 0.5) -- (\i+0.45, 0.65) -- (\i+0.45, -0.35) -- cycle;
    }
\end{scope}

\draw[arrow] (add2) -- (1.2, -8.6);

\end{tikzpicture} }
\vspace{0.4cm}
\caption{Channel-wise attention module.}
\label{fig:cahnnel}
\end{figure}

First, a global average pooling operation is applied across the spatial dimensions of $\mathbf{F}$ to produce a channel descriptor resulting in a compact vector $\mathbf{z} \in \mathbb{R}^{C}$ representing the global channel statistics. To incorporate channel adaptivity, the SNR value is also processed through a multi layer perceptron (MLP) to generate a vector of arbitrary size, which is usually smaller than number of channels $C$. The two vectors are then concatenated and passed through a small MLP consisting of two fully connected layers separated by a parametric rectified linear unit (PReLU) activation, followed by a sigmoid activation to generate the adaptive channel attention weights
\begin{equation}
    \mathbf{w} = \sigma_1\left({\rm MLP}\left[\sigma_2\left({\rm MLP}\left({\rm SNR}\right)\right)^T,\mathbf{z}^T\right]^T\right).
\end{equation}
Here, $\sigma_1(\cdot)$ and $\sigma_2(\cdot)$ represent the corresponding activation functions depicted in Fig. \ref{fig:cahnnel}. The resulting vector $\mathbf{w} \in \mathbb{R}^{C \times 1}$ contains normalized weights between 0 and 1 that represent the importance of each channel.
Finally, the attention-weighted feature map is obtained by performing an element-wise multiplication between the input feature map and the channel weights, followed by a residual addition to preserve the original feature information:
\begin{equation}
    \mathbf{F}_{{\rm out}} = \mathbf{F} + \mathbf{w} \otimes \mathbf{F},
\end{equation}
where $\otimes$ denotes channel-wise multiplication.

The adaptive spatial attention module focuses on identifying the most informative spatial regions within the feature map. Given an input feature map $\mathbf{F} \in \mathbb{R}^{H \times W \times C}$, the module generates a spatial attention map that emphasizes semantically important locations while being robust to channel variations.

First, spatial information is aggregated by performing global average pooling across the channel dimension:
\begin{equation}
    \mathbf{F}_{{\rm avg}} = \frac{1}{C} \sum_{c=1}^{C} \mathbf{F}_c,
\end{equation}
where $\mathbf{F}_c$ denotes the feature map of channel $c$, and  $\mathbf{F}_{{\rm avg}} \in \mathbb{R}^{H \times W}$ represents the average feature response across all channels. To incorporate channel adaptivity, the SNR value is expanded and reshaped into a feature map 
$\mathbf{F}_{{\rm snr}} \in \mathbb{R}^{H \times W}$ by broadcasting or learned upsampling. This allows the network to integrate channel condition awareness into spatial attention module.

The two maps are concatenated along the channel dimension and passed through a convolutional layer to learn the spatial attention weights:
\begin{equation}
    \mathbf{M}_s(\mathbf{F}) = \sigma_3\left({\rm Conv}_{k \times k}\left[\mathbf{F}_{{\rm avg}},\mathbf{F}_{{\rm snr}}\right]\right),
\end{equation}
where $\mathbf{M}_s(\mathbf{F}) \in \mathbb{R}^{H \times W}$ is the spatial attention map and $\sigma_3(\cdot)$ denotes the sigmoid activation function.

\begin{figure}[t!]
\centering
\resizebox{0.25\textwidth}{!}{
\begin{tikzpicture}[
    scale=0.85,
    transform shape,
    font=\sffamily\small\bfseries,
    every node/.style={align=center},
    box/.style={
        draw,
        thick,
        minimum width=1.4cm,
        minimum height=0.6cm,
        fill opacity=0.85,
        text opacity=1,
        rounded corners=2pt
    },
    op/.style={
        circle,
        draw,
        thick,
        minimum size=0.4cm
    },
    arrow/.style={-latex, thick}
]


    \node[box, fill=orange!35] (input) {Input \\ $\mathbf{F}$};
    \node[box, fill=blue!25, below=0.6cm of input] (avg) {AvgPool};
    \node[box, fill=cyan!25, right=1.0cm of avg] (snr) {SNR};
    \node[box, fill=purple!55, below=0.5cm of avg] (add1) {Concatenator};
    \node[box, fill=gray!25, below=0.6cm of add1] (conv) {Conv2D};
    \node[box, fill=green!30, below=0.6cm of conv] (attn) {Spatial\\Weights};
    \node[box, fill=purple!25, below=0.6cm of attn] (sigmoid) {Sigmoid};
    \node[op, below=0.6cm of sigmoid] (mult) {$\times$};
    \node[op, below=0.6cm of mult] (add2) {\small +};
    \node[box, fill=orange!35, below=0.6cm of add2] (output) {Output \\ $\mathbf{F}_{\rm out}$};

    \draw[arrow] (input.south) -- (avg.north);       
    \draw[arrow] (avg.south) -- (add1.north);        
    \draw[arrow] (snr.south) |- (add1.east);         
    \draw[arrow] (add1.south) -- (conv.north);       
    \draw[arrow] (conv.south) -- (attn.north);       
    \draw[arrow] (attn.south) -- (sigmoid.north);    
    \draw[arrow] (sigmoid.south) -- (mult.north);    
    \draw[arrow] (mult.south) -- (add2.north);       
    \draw[arrow] (add2.south) -- (output.north);     

    \draw[arrow]
        (input.south west) ++(-0.1,-0.1)
        .. controls +(-1.2,-1.0) and +(-1.0,0.5) ..
        (mult.west);
    \draw[arrow]
        (input.south east) ++(0.1,-0.1)
        .. controls +(1.2,-1.0) and +(1.0,0.5) ..
        (add2.east);

    \node[
        draw=gray!60,
        dashed,
        rounded corners=6pt,
        fit=(input)(output),
        inner sep=15pt
    ] {};

    \end{tikzpicture} }

    \caption{Spatial attention module.}
    \label{fig:spatial_attention}
\end{figure}
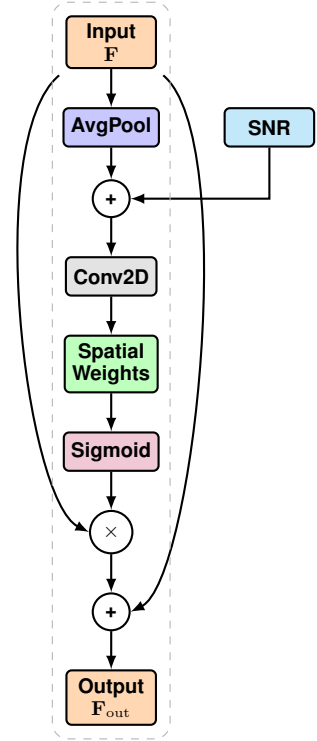

The resulting attention weights highlight spatially significant regions of the feature map. The attention-weighted feature map is then obtained through element-wise multiplication, followed by a residual addition to preserve the original feature information:
\begin{equation}
    \mathbf{F}_{{\rm out}} = \mathbf{F} + \mathbf{F} \odot \mathbf{M}_s(\mathbf{F}),
\end{equation}
where $\odot$ denotes element-wise multiplication. Please also check Fig. \ref{fig:spatial_attention} for complete details.

Through this mechanism, the spatial attention module enhances semantic regions in the feature map while maintaining robustness to noise and other distortions. This improves the effectiveness of semantic feature reconstruction under dynamic wireless channel conditions.

\section{Numerical Results}
First, we describe the implementation setup, training configurations, and datasets used for experimentation. Then, we present and analyze the quantitative results, where the proposed approach is compared against available benchmarks under varying SNR conditions. These experiments demonstrate the robustness, efficiency, and adaptability of our proposed approach compared to existing alternatives.

\subsection{Training and Implementation Details}
The proposed model was evaluated on two benchmark datasets: CIFAR-10 and CIFAR-100. Both datasets contain 50,000 training images and 10,000 test images of size $32 \times 32 \times 3$. CIFAR-10 has 10 classes, while CIFAR-100 has 100 classes. The model was trained with a batch size of 32 for 300 epochs. We employed the Adam optimizer with learning rate set to $\text{lr} = 1 \times 10^{-4}$, and weight decay set to $1 \times 10^{-5}$. Mean squared error (MSE) loss function is used to optimize the reconstruction quality between the original and reconstructed images.

Three key performance metrics are plotted to demonstrate improved performance of the proposed approach. These metrics are structural similarity index measure (SSIM), peak signal-to-noise ratio (PSNR), and classification accuracy. PSNR pertains to reconstruction fidelity, and measures the ability to improve pixel-level reconstruction quality as channel conditions improve. Superior SSIM scores indicate that the architecture excels at preserving the structural and semantic information essential for human perception and machine analysis. Finally, classification accuracy measures the impact of semantic communications on the performance of the downstream machine learning task. Together, these three metrics provide a balanced evaluation of both the signal-level reconstruction quality (SSIM and PSNR) and the task-level semantic fidelity (Accuracy), ensuring that model performs effectively under diverse wireless channel conditions.

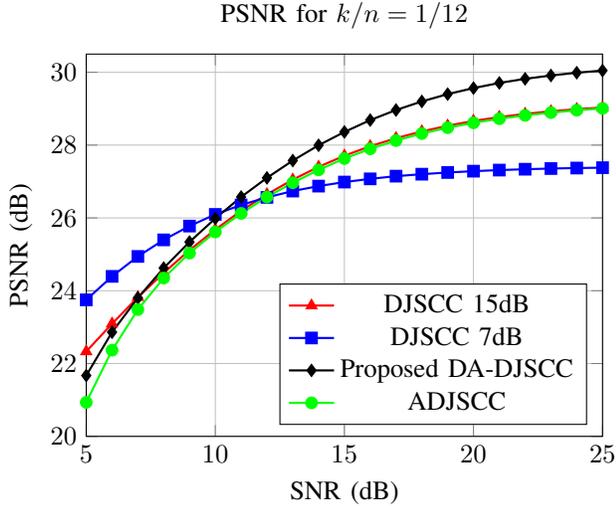
\begin{figure}[t!]
\centering
\begin{tikzpicture}
\begin{axis}[
    width=0.475\textwidth,
    height=0.375\textwidth,
    xlabel={SNR (dB)},
    ylabel={PSNR (dB)},
    xmin=5, xmax=25,
    ymin=20, ymax=30.5,
    grid=major,
    grid style={line width=.1pt, draw=gray!30},
    major grid style={line width=.2pt,draw=gray!50},
    mark size=2pt,
    legend pos=south east,
]
\addplot[color=red, mark=triangle*, thick]
    table[col sep=comma, x=SNR_dB, y=PSNR_dB] {snr_evaluation_15.csv};

\addplot[color=blue, mark=square*, thick]
    table[col sep=comma, x=SNR_dB, y=PSNR_dB] {snr_7_dB.csv};

\addplot[color=black, mark=diamond*, thick]
    table[col sep=comma, x=SNR_dB, y=PSNR_dB] {snr_evaluation_CODE_4.csv};

\addplot[color=green, mark=*, thick]
    table[col sep=comma, x=SNR_dB, y=PSNR_dB] {psnr_ssim_vs_snr_w.csv};

\legend{DJSCC 15dB,DJSCC 7dB, Proposed DA-DJSCC, ADJSCC \cite{b7}}
\end{axis}
\end{tikzpicture}
\caption{PSNR performance for $k/n=1/12$.}
\label{fig:psnr_kn}
\end{figure}

\begin{figure}[t!]
\centering
\begin{tikzpicture}
\begin{axis}[
    width=0.475\textwidth,
    height=0.375\textwidth,
    xlabel={SNR (dB)},
    ylabel={SSIM},
    xmin=5, xmax=25,
    ymin=0.7, ymax=0.95,
    grid=major,
    grid style={line width=.1pt, draw=gray!30},
    major grid style={line width=.2pt,draw=gray!50},
    mark size=2pt,
    legend pos=south east,
]
\addplot[color=red, mark=triangle*, thick]
    table[col sep=comma, x=SNR_dB, y=SSIM] {snr_evaluation_15.csv};

\addplot[color=blue, mark=square*, thick]
    table[col sep=comma, x=SNR_dB, y=SSIM] {snr_7_dB.csv};

\addplot[color=black, mark=diamond*, thick]
    table[col sep=comma, x=SNR_dB, y=SSIM] {snr_evaluation_CODE_4.csv};

\addplot[color=green, mark=*, thick]
    table[col sep=comma, x=SNR_dB, y=SSIM] {psnr_ssim_vs_snr_w.csv};

\legend{DJSCC 15dB,DJSCC 7dB, Proposed DA-DJSCC, ADJSCC \cite{b7}}
\end{axis}
\end{tikzpicture}
\caption{SSIM performance for $k/n=1/12$.}
\label{fig:ssim_kn}
\end{figure}
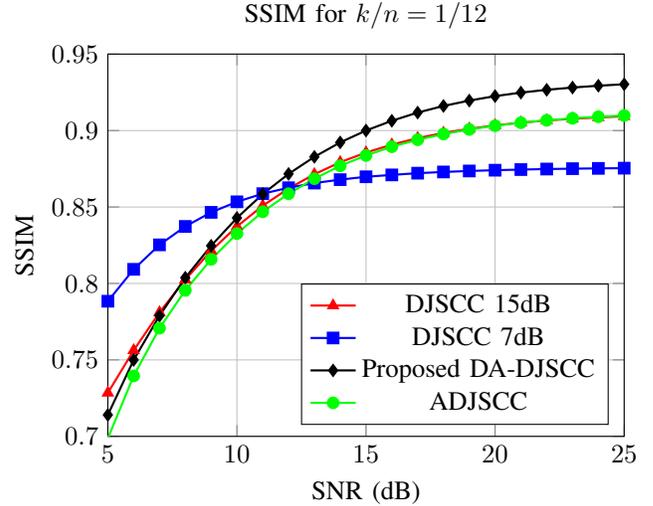

\begin{figure}[t!]
\centering
\begin{tikzpicture}
\begin{axis}[
    width=0.475\textwidth,
    height=0.375\textwidth,
    xlabel={SNR (dB)},
    ylabel={Accuracy (\%)},
    xmin=5, xmax=25,
    ymin=40, ymax=94,
    grid=major,
    grid style={line width=.1pt, draw=gray!30},
    major grid style={line width=.2pt,draw=gray!50},
    mark size=2pt,
    legend pos=south east,
]
\addplot[color=red, mark=triangle*, thick]
    table[col sep=comma, x=SNR_dB, y=Accuracy_Percent] {snr_evaluation_15.csv};

\addplot[color=blue, mark=square*, thick]
    table[col sep=comma, x=SNR_dB, y=Accuracy_Percent] {snr_7_dB.csv};

\addplot[color=black, mark=diamond*, thick]
    table[col sep=comma, x=SNR_dB, y=Accuracy_Percent] {snr_evaluation_CODE_4.csv};

\addplot[color=green, mark=*, thick]
    table[col sep=comma, x=SNR_dB, y=Accuracy_Percent] {psnr_ssim_vs_snr_w.csv};

\legend{DJSCC 15dB,DJSCC 7dB, Proposed DA-DJSCC, ADJSCC \cite{b7}}
\end{axis}
\end{tikzpicture}
\caption{Classification accuracy for $k/n=1/12$.}
\label{fig:acc_kn}
\end{figure}
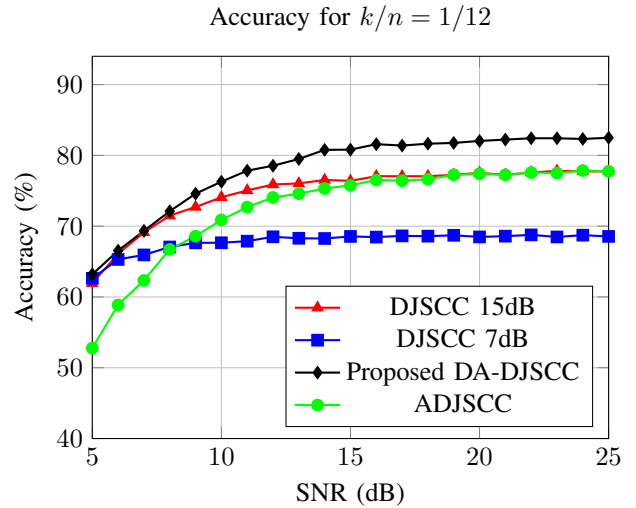

\renewcommand{\arraystretch}{1.5}
\begin{table}[t]
\centering
\caption{Comparison of model complexity for proposed model and baseline DJSCC models.}
\label{tab:model_complexity}
\resizebox{0.45\textwidth}{!}{%
\begin{tabular}{|c|c|c|c|}
\hline
\textbf{Model} & \multicolumn{3}{|c|}{\textbf{Complexity Metrics}} \\
\cline{2-4} 
& \textbf{Parameters} & \textbf{Memory (MB)} & \textbf{Latency (ms)} \\
\hline
DJSCC   & 0.143M & 0.547 & 0.5 \\
\hline
ADJSCC  & 0.157M & 0.601 & 2 \\
\hline
Proposed DA-DJSCC & 0.159M & 0.607 & 3 \\
\hline
\multicolumn{4}{l}{$^{\mathrm{a}}$All values are measured under the same hardware and batch size conditions.}
\end{tabular}%
}
\end{table}

\subsection{Results and Analysis}

For comparison, two benchmarks are selected. First one is the original DJSCC approach proposed by \cite{b6}, which includes neither attention modules nor SNR adaptive mechanisms. We have trained DJSCC under two SNR values of $7$ and $15$ dB, enabling evaluation under a variety of channel conditions. The second benchmark is the SNR adaptive channel-wise attention mechanism proposed by \cite{b7}, which is the ADJSCC. This benchmark lacks the spatial attention module proposed in this work. It needs to be mentioned that the provided comparisons can be considered as an ablation study of the proposed algorithm as well. Indeed, DJSCC lacks SNR adaptivity in both channel and spatial dimensions, while ADJSCC incorporates SNR adaptivity only in the channel dimension. Finally, our proposed DA-DJSCC incorporates SNR adaptivity in both channel and spatial dimensions. 

Figs. \ref{fig:psnr_kn}, \ref{fig:ssim_kn}, and \ref{fig:acc_kn} illustrate the performance of the proposed DA-DJSCC in terms of PSNR, SSIM, and classification accuracy, respectively. The benchmarks are also plotted for comparison. All results are reported for a fixed compression ratio of $k/n=1/12$. It is worth noting that classification accuracy results are obtained from models trained on CIFAR-10, whereas SSIM and PSNR evaluations are conducted using models trained on the CIFAR-100 dataset. In Figs. \ref{fig:psnr_kn} and \ref{fig:ssim_kn}, it can be observed that the proposed DA-DJSCC outperforms the benchmarks over a wide range of SNR values. For small SNR values, the benchmark DJSCC trained with an SNR of $7$ dB outperforms other methods. This can be attributed to the fact that with large channel noises, stronger models' parameters tend to fit the noise rather than the desired image. Hence, at low SNRs a simple DJSCC, which has less parameters to tune, outperforms other approaches. Above the $10$ dB SNR our proposed DA-DJSCC offers a significant improvement. The ADJSCC benchmark performs poorly at low SNRs, which is expected given the attention mechanism and possible overfitting. However, ADJSCC also improves with SNR and outperforms the DJSCC benchmark. Notably, it still performs worse than our proposed method with a significant margin, which is due to the lack of a spatial attention module in ADJSCC.

Fig. \ref{fig:acc_kn} demonstrates the performance of our proposed approach for a downstream classification task at the receiver or the machine learning server connected to it. Classification accuracy evaluates the ability of the model to preserve task-relevant semantic information, which is essential for intelligent IoT applications such as sensing, recognition, and decision-making. For this evaluation, we used a pretrained VGG16 classifier on the CIFAR-10 dataset which achieves a maximum baseline accuracy of 94\% on clean images. It can be observed that our proposed method outperforms all benchmarks in classification accuracy over the entire SNR range.

Finally, computational complexity of various approaches are provided in Table I. Specifically, the table reports the number of trainable parameters, memory required for model storage, and inference latency, providing a comprehensive assessment of both storage and computational efficiency. It can be observed that our proposed DA-DJSCC is more complex than the two benchmarks. For example, it is 50\% more complex than ADJSCC. While our complexity is higher, it is not significantly higher and it is in the same order of the two benchmarks. As advantages, our DA-DJSCC does not suffer from the required weight modification of DJSCC at various SNRs, which incurs a large storage and communication complexity. This can be explained by the fact that DA-DJSCC requires 0.607 MB of storage for the entire SNR range, while DJSCC requires 0.547 MB per training SNR. Even for a very simple case of two training SNRs, DJSCC demands about 1.1 MB of storage which is significantly larger than DA-DJSCC. DA-DJSCC also outperforms ADJSCC in all three metrics over the entire SNR range. Hence, it offers a good complexity-performance trade-off for IoT networks.

\section{Conclusion}

This work presented a novel doubly adaptive semantic communication architecture for IoT applications, designed to be robust under varying and noisy channel conditions. The core theme of the proposed approach lies in the doubly adaptive attention modules that dynamically recalibrate spatial and channel-wise features based on SNR. This allows the model to prioritize the most informative features for efficient extraction and robust recovery.

Experimental evaluations demonstrated the superiority of the proposed framework over conventional DJSCC benchmarks across three critical metrics, which were selected as PSNR, SSIM, and classification accuracy of the downstream machine learning task. Numerical results confirm that the proposed doubly adaptive semantic communication architecture is robust to channel conditions across a wide range of SNR by dynamically aligning the encoding process with the physical channel state. Thus, this work provides a significant step toward reliable and efficient semantic communication for the next-generation IoT systems.

\end{document}